\documentclass[10pt,twocolumn,letterpaper]{article}

\usepackage{cvpr}              
\usepackage{soul}
\usepackage{float}
\usepackage{cuted}
\usepackage{multirow} 

\definecolor{cvprblue}{rgb}{0.21,0.49,0.74}
\usepackage[pagebackref,breaklinks,colorlinks,allcolors=cvprblue]{hyperref}

\def\confName{CVPR}
\def\confYear{2025}
\usepackage{hyperref}
\title{A Plug-and-Play Physical Motion Restoration Approach \\
for In-the-Wild High-Difficulty Motions}

\author{
    Youliang Zhang$^{1}$\thanks{Equal contribution}\quad
    Ronghui Li$^{1}$\footnotemark[1]\quad  
    Yachao Zhang$^1$\quad \\
    Liang Pan$^2$\quad Jingbo Wang$^2$\quad Yebin Liu$^1$\quad 
    Xiu Li$^{1}$\thanks{Corresponding author} \\
    \normalsize{${^1}$Tsinghua University, China} \quad
    \normalsize{${^2}$Shanghai AI Laboratory}
}

\begin{document}

\maketitle
\begin{strip}
\centering
    \vspace{-50pt}
    \includegraphics[width=\textwidth]{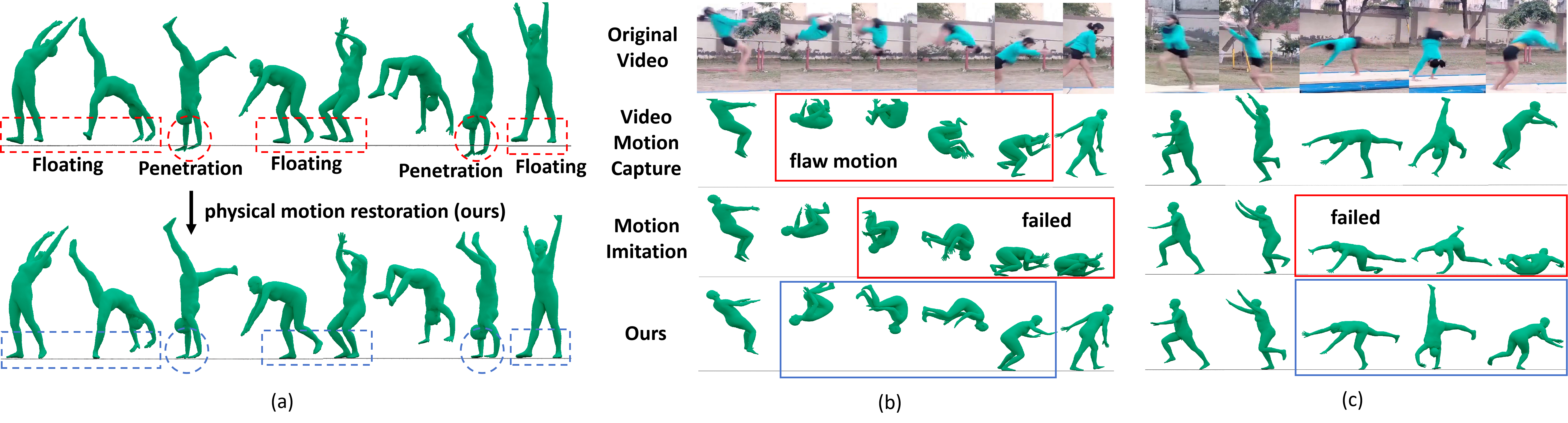}
    \vspace{-24pt}
    \captionof{figure}{\textbf{Illustration of motivation and two main challenges.} 
    (a) Our method effectively enhances the physical plausibility of video-captured motions, even handling high-difficulty motions like backflips.
    (b) highlights the challenging movements in the original video lead to flawed motion estimated by current video motion capture algorithms, 
    where the current motion imitation model fails to restore overly degraded flawed motions.
    (c) demonstrates that even when video motion capture provides reasonable reference motions, existing motion imitation techniques still fail to track complex motions.}
    \label{fig:motivation}
\end{strip}
\vspace{-20pt}
\begin{abstract}
Extracting physically plausible 3D human motion from videos is a critical task. 
Although existing simulation-based motion imitation methods can enhance the physical quality of daily motions estimated from monocular video capture, extending this capability to high-difficulty motions remains an open challenge. This can be attributed to some flawed motion clips in video-based motion capture results and the inherent complexity in modeling high-difficulty motions.
Therefore, sensing the advantage of segmentation in localizing human body, we introduce a mask-based motion correction module (MCM) that leverages motion context and video mask to repair flawed motions, producing imitation-friendly motions; 
and propose a physics-based motion transfer module (PTM), which employs a pretrain and adapt approach for motion imitation, improving physical plausibility with the ability to handle in-the-wild and challenging motions.
Our approach is designed as a plug-and-play module to physically refine the video motion capture results, including high-difficulty in-the-wild motions.
Finally, to validate our approach, we collected a challenging in-the-wild test set to establish a benchmark, and our method has demonstrated effectiveness on both the new benchmark and existing public datasets. \href{https://physicalmotionrestoration.github.io/}{https://physicalmotionrestoration.github.io/}

\vspace{-10pt}
\end{abstract}

\vspace{-10pt}
\section{Introduction}
\label{sec:intro}



Physical plausible 3D human motion is in high demand across various fields, including virtual reality, game, animation industries, and academic research on virtual humans and robotics \cite{colombel2020physically,erickson2020assistive,mehta2017vnect,wang2024freeman,grauman2024ego,rempe2023trace,wang2023learning,LiuZZJWL23,zhang2024bi,luo2022embodied}. 
With technological advancements, monocular video-based motion capture algorithms provide a fast and convenient pipeline for obtaining 3D motion closely aligned with video.
However, these methods inherently lack dynamic modeling, resulting in significant physical unrealisms such as floating, foot sliding, self-intersections, and ground penetration, things get worse when facing high-difficulty motions.

To enhance physical realism, some methods use dynamic equations and train a network to predict physical parameters \cite{tripathi20233d,li2022d,gartner2022trajectory,xie2021physics}. However, these methods often struggle to improve motion plausibility due to oversimplified dynamic equations.
Other methods \cite{yuan2021simpoe,shimada2020physcap,gong2022posetriplet,gartner2022differentiable} use physical-simulation-based motion imitation as a post-processing module, learning motion control policy to imitate the reference motion (video motion capture results) in a simulated physical environment.
With high-quality reference motions, these methods improve the physical realism of daily motions such as walking, running, and jumping. However, they can not handle high-difficulty movements like gymnastics and martial arts.
Regarding this, we aim to extend the physical restoration ability of motion imitation in high-difficulty and in-the-wild motions, meeting broader requirements for motion asset acquisition.

Reviewing the characteristics of high-difficulty motions, they often involve rapid movement, extreme poses, skilled force control, and follow a long-tail distribution in existing datasets. 
This presents two major challenges for existing motion imitation methods to enhance the physical plausibility of complex motions within a physical simulation environment:
(1) \textbf{Flawed Reference Motions}: As shown in Fig~\ref{fig:motivation}(b), even the state-of-the-art video motion capture algorithms estimate flawed motions when facing challenging movements. 
Such brief disruptions can easily cause failures in the motion imitation process and are obvious in the human senses.
(2) \textbf{Inherent Imitation Complexity}: The long-tail distribution of high-difficulty motions and their complex force control skills make it challenging for current motion imitation methods to track high-difficulty motions, 
shown in Fig~\ref{fig:motivation}(c).
Moreover, a single controller struggles to generalize across a diverse range of high-difficulty movements, facing catastrophic forgetting issues, where rapid loss of old knowledge occurs when learning new skills \cite{luo2023perpetual}.

To solve the issue of \textbf{flawed reference motion}, we propose a mask-conditioned correction module (MCM). 
Due to the blurred frames caused by rapid and extreme poses, video motion capture algorithms struggle to localize body parts accurately when facing high-difficulty motions. 
Notably, the nature of segmentation methods to distinguish the foreground and background mitigates the effects of blurred frames, allowing them to define an approximate range of body.
Also, the flaw motion occurs over a short time and is surrounded by rich motion contexts, making segmentation-guided interpolation and replacement of flaw motion possible.
We first leverage a large visual semantic segmentation model \cite{kirillov2023segany} to obtain human segmentation masks of the input video. Then, guided by segmentation masks and reference motion context, we regenerate context-consistent and imitation-friendly motions to replace the flawed motion, allowing accurate and complete motion imitation.

To tackle the \textbf{inherent imitation complexity} of diverse challenging movements, we propose a Physics-based Motion Transfer Module (PTM) consisting of a pre-trained imitation controller and a test time adaptation strategy (TTA). 
Utilizing the trial-and-error nature of reinforcement learning and pre-trained motion prior, we targeted adapting our network to current motion in test time, naturally addressing the long-tail distribution and domain gap issues by parameter updating. 
To better achieve complex force control, our TTA contains unique imitation settings for adaptation to facilitate tracking the noisy yet challenging motions captured from videos.
The pretrain and adapt pattern of PTM greatly improves the imitation ability, allowing the successful physical restoration of in-the-wild and high-difficulty motions. 

Through our proposed novel MCM and PTM, we successfully address the failure issues of flaw motion and complex motion simulation, achieving physical authenticity restoration in high-difficulty motions while faithfully retaining the original movements. It is worth mentioning that without additional training, our method can be conveniently integrated into any video motion capture method, directly repairing in-the-wild and high-difficulty motions.

To validate the effectiveness of our proposed motion restoration method, we collected 206 high-difficulty motion videos entirely in the wild, including activities such as rhythmic gymnastics, taekwondo, and yoga. 
Our method demonstrated strong performance on this in-the-wild test set, which is significantly more challenging than the training set, further proving the effectiveness of our approach.

Our contributions can be summarized as follows:
\begin{itemize}
    \item We propose a plug-and-play motion restoration method that enhances the physical realism of motions captured from monocular video. It can be integrated with existing motion capture algorithms to improve the efficiency of obtaining high-quality 3D motion.

    \item We introduced an MCM for correcting short-term flawed motions, producing consistent and imitation-friendly motions for physical restoration.
    
    \item We introduced a PTM with a pretrain and adapt motion imitation pattern, allowing the physical restoration of high-difficulty and in-the-wild motions.

    \item We collected a challenging in-the-wild test set to establish a benchmark, and our method demonstrates effectiveness on both the new benchmark and existing public datasets.
\end{itemize}



\section{Related Work} 
\label{sec:related}

\begin{figure*}[t]
\vspace{-25pt}
    \centering
\includegraphics[width=0.8\textwidth, height=0.28\textwidth]{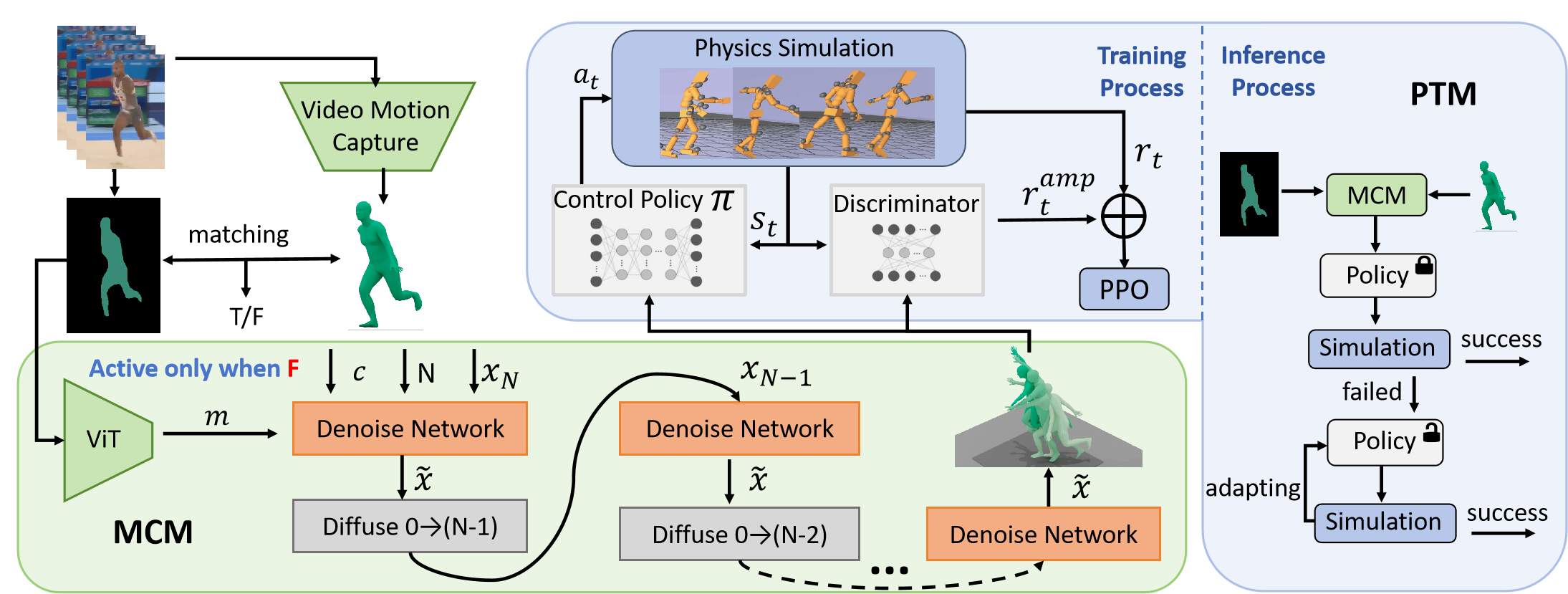}
     \vspace{-2mm}
    \caption{\label{fig:pipeline}\textbf{Illustration of our proposed method.} If no mismatch is detected between the human mask and noise motion, the correction process will be skipped, and our PTM directly takes the noise motion as input.
    When failed with challenge motions, our PTM will adapt the policy to the current motion and update the network parameters until success or reach a certain step threshold. 
    }
     \vspace{-5mm}
\end{figure*}

\subsection{Video Motion Capture}
Most works for video motion capture are to recover the parameters of a parametric human model \cite{loper2023smpl,kanazawa2019learning,choi2021beyond, kocabas2020vibe,luo20203d,sun2019human,shen2023global,tian2023recovering,dwivedi2024tokenhmr,ma2023grammar,ge20243d}. 
Recently, many methods have started to consider moving cameras.
TRACE \cite{sun2023trace} and WHAM \cite{shin2024wham} propose regressing per-frame poses and translations.
SLAHMR \cite{ye2023decoupling} and PACE \cite{kocabas2024pace} integrate SLAM \cite{teed2021droid,teed2024deep} with motion priors \cite{rempe2021humor} into the optimization framework.
TRAM \cite{wang2025tram} leverages the scene background to derive motion scale.
GVHMR \cite{shen2024world} estimates human poses in a novel Gravity-View coordinate system.
While these methods achieve significant success in reconstructing high-difficulty motions from videos, they suffer from serious physical issues and occasionally experience flawed motions when facing complex movements.
Our proposed physical motion restoration method effectively addresses these problems.

\subsection{Motion Imitation}
The physical constraints provided by simulation environments give simulated characters a clear advantage in generating lifelike human movements \cite{peng2018deepmimic,peng2017deeploco,peng2019mcp,peng2022ase,peng2021amp,chentanez2018physics,merel2020catch,winkler2022questsim,gong2022posetriplet,fussell2021supertrack,wang2024skillmimic,luo2024smplolympics,wang2024pacer+}.
Early works focused on small-scale task-specific scenarios and are difficult to generalize to other domains.
With the advancements in motion generation technology \cite{lu2023humantomato}, training policies to imitate large-scale motion datasets show broader application potential \cite{peng2022ase}. 
Researchers improve motion simulation quality by leveraging techniques such as hybrid expert policies \cite{won2020scalable}, differentiable simulation \cite{ren2023diffmimic}, and external forces \cite{yuan2020residual}.
ScaDiver \cite{won2020scalable} extended the hybrid expert strategy to the CMU motion capture dataset. Unicon \cite{wang2020unicon} demonstrated qualitative results in imitation and transfer tasks.
MoCapAct \cite{wagener2022mocapact} learn a single-segment expert policy on the CMU dataset. 
UHC \cite{luo2021dynamics} successfully imitated 97\% of the AMASS dataset, and recently, PHC \cite{luo2023perpetual, luo2023universal} enabled a single policy to simulate almost the entire AMASS dataset while allowing recovery from falls.
However, these methods rely heavily on the quality of reference motions and are largely confined to locomotion tasks. 
Simulating in-the-wild and high-difficulty motions is still challenging, where our proposed PTM provides a possible solution.

\subsection{Physics informed Video Motion Capture}
Many researchers attempt to introduce physics into video motion capture.
Some methods \cite{shimada2021neural,huang2022neural,gartner2022trajectory,tripathi20233d, le2024optimal} leverage neural networks to estimate physical parameters for motion capture and introduce the kinematic constrains to enhance the physical plausibility. LEMO \cite{zhang2021learning} uses motion smoothness prior and physics contact friction term. Xie ~\etal \cite{xie2021physics} propose differentiable physics-inspired objectives with contact penalty. IPMAN \cite{tripathi20233d} exploits intuitive-physics terms to incorporate physics. Li ~\etal \cite{li2022d} enhanced the learning process by incorporating 3D supervision. 
These methods typically require hard-to-obtain 3D annotations and overly simplify dynamic equations, struggling with generalization to out-of-distribution motions.
There are also methods combine motion imitation to enhance the physical plausibility, they treat the captured motion as a reference and predict the physical simulation forces with a controller \cite{yuan2021simpoe,gong2022posetriplet,zhang2024proxycap}. 
DiffPhy \cite{gartner2022differentiable} uses a differentiable physics simulator during inference. 
PhysCap \cite{shimada2020physcap} uses a numerical optimization framework with soft physical constraints.
SimPoE \cite{yuan2021simpoe} integrates image-based kinematic inference and physics-based dynamics modeling.
However, these methods typically require careful tuning of control parameters and are sensitive to different motion types \cite{huang2022neural}. 
This sensitivity makes it challenging to generalize to in-the-wild high-difficulty motions, limiting real-world applications.
Recently, PhysPT \cite{zhang2024physpt} proposed a pre-trained physics-aware transformer to learn human dynamics in a self-supervised manner. 
However, it lacks an understanding of the distribution and physical rules of high-difficulty motions, necessitating additional physical priors of complex motions, which is challenging due to data scarcity. 
In contrast, our approach is designed to restore high-difficulty and in-the-wild motions while maintaining their original motion pattern.

\vspace{-7pt}
\section{Physics-based Motion Restoration}
Our method takes video-captured motion as reference motions and focuses on restoring their physical realism while preserving original motion patterns.
The motion representation $\boldsymbol{x}_t$ consists of joint position $\boldsymbol{p}_t \in \mathbb{R}^{J \times 3}$ and rotation $\boldsymbol{\mathcal{\theta}}_t \in \mathbb{R}^{J \times 6}$ \cite{zhou2019continuity}, compatible with SMPL format \cite{loper2023smpl}. $J$ means the joint number of the humanoid.
The reference velocities $\boldsymbol{q}_t$ is also considered, which consists of the linear $\boldsymbol{v}_t \in \mathbb{R}^{J \times 3}$ and angular $\boldsymbol{\omega}_t \in \mathbb{R}^{J \times 6}$ velocity. 
An overview of our method is provided in Fig \ref{fig:pipeline}. 
Given the reference motion (video motion capture results) and corresponding video, our MCM detects and corrects the flawed motion. Our PTM inputs the corrected motion and performs physical restoration by motion imitation. In PTM, a pre-trained controller and a carefully designed adaptation strategy are well-cooperated to solve the dynamics of a single motion. This pre-train and adapt pattern makes our PTM perform well in tracking high-difficulty and in-the-wild motions.


\subsection{Preliminaries}
\textbf{Motion Imitation.}
The problem of controlling a humanoid to follow a reference motion sequence can be formulated as a Markov Decision Process, defined by the tuple $M = \langle S, A, P_{\text{physics}}, R, \gamma \rangle$, which consists of states, actions, transition dynamics, reward, and a discount factor. 
At step $t$, agent samples an action $\boldsymbol{a}_t$ from the policy $\pi_{\text{PTM}}(\boldsymbol{a}_t|\boldsymbol{s}_t)$ based on the current state $\boldsymbol{s}_t$, and the environment responds with the next state $\boldsymbol{s}_{t+1}$ and a reward $r_t$. Proximal Policy Optimization \cite{schulman2017proximal} is used to optimize the policy $\pi^*_{\text{PTM}}$ by maximize the expected discounted return $\mathbb{E}\left[\sum_{t=1}^{T}\gamma^{t-1}r_{t}\right]$.
The {state} $\boldsymbol{s}_{t}$ consists of positions, rotations, and linear and angular velocities of humanoid, as well as the next frame information $\boldsymbol{g}_t$.  We define $\boldsymbol{g}_t$ as the difference between the current frame and next frame reference motion \cite{yuan2023physdiff, luo2023perpetual}.
The {action} $\boldsymbol{a}_t$ specifies the target humanoid joint angles for the controller at each degree of freedom (DoF).
Given an action and current motion $\boldsymbol{x}_t$ and velocity $\boldsymbol{q}_t$, the torque to be applied is computed as:
\begin{equation}
\boldsymbol{\tau}^i=\boldsymbol{k}_p^i(\boldsymbol{a}_t^i-\boldsymbol{x}_t^i)-\boldsymbol{k}_d^i\boldsymbol{q}_t^i ,
\end{equation}
where $i$ means joint DoF index, $\boldsymbol{k}_p$ and $\boldsymbol{k}_d$ are manually-specified gains. 
Our {policy} $\pi_{\text{PTM}}$ is constructed with multilayer perceptions and ReLU functions. A discriminator from AMP \cite{peng2021amp} is used to predict whether a given state $\boldsymbol{s}_t$ and action $\boldsymbol{a}_t$ is sampled from the demonstrations $M$ or generated by policy $\pi_{\text{PTM}}$. 
The {reward} consists of a reconstruction reward $r_t^\mathrm{g}$ to follow the reference motion, a style reward $r_t^\mathrm{amp}$ produced by the amp discriminator, and an energy penalty reward $r_t^\mathrm{energy}$ \cite{peng2018deepmimic} to prevents motion jitter. 
\begin{equation}r_t=r_t^\mathrm{g}+r_t^\mathrm{amp}+r_t^\mathrm{energy}.\end{equation}

\textbf{Motion Diffusion Model.}
The diffusion model consists of a forward diffusion process that progressively adds noise to the clean data and a reverse diffusion process trained to reverse this process. The forward diffusion process introduces noise for $N$ steps formulated using a Markov chain:
\begin{equation}q(\boldsymbol{x}_{1:N}|\boldsymbol{x}_0):=\prod_{n=1}^Nq(\boldsymbol{x}_n|\boldsymbol{x}_{n-1}),\end{equation}
Reverse process employs a learnable network $f_{\theta}$ to denoise. 
\subsection{Mask-conditioned Motion Correction Module}
Flaw motions in video capture results manifest as incoherent frames within the motion sequence inconsistent with the surrounding motion context. Although such brief disruptions have a minimal impact on metrics, they can easily cause failures in the physics simulation process and are obvious in the human senses.
To address this issue, our MCM first detects flawed motion and then regenerates the flawed motion segment guided by the motion context and human mask signals, ultimately replacing the flawed motion.

\textbf{Mismatch Detection.} 
Given the reference motion and its corresponding video, we project the 3D positions of the reference motion into 2D camera coordinates. Also, an object detection algorithm is used to extract the corresponding 2D keypoints from the video. Using the Object Keypoint Similarity (OKS) algorithm, we compute the matching degree between the two sets of keypoints and obtain a similarity score sequence. Frames with a similarity score below a certain threshold will be flagged as flaw motion. 
\begin{equation}OKS=\frac{\sum_iexp(-d_i^2/2\epsilon_i^2)\delta(v_i>0)}{\sum_i\delta(v_i>0)}\end{equation}
where $v_i$ represents the visibility flag, $\epsilon_i$ denotes the scale factor, and $d_i$ is the distance difference between the projection and the detection results.
Additionally, segmentation algorithms can also be used to detect flaw motion. We project the SMPL-generated mesh onto the 2D plane and treat it as a set of pixel points. 
The matching similarity can be calculated by determining the proportion of projected mesh points that are contained within the human mask.


\textbf{Motion Correction Goal Design.}
Unlike traditional motion in-between methods, the goal of  MCM is to replace flawed motions in the reference sequence based on contextual information, ensuring that the corrected motion is smooth and reasonable. Thus, we focus on the model’s ability to fill temporal gaps. 
Since the input is motion captured from video, ensuring the generated motion remains consistent with the original video content is crucial. Therefore, we use the human mask obtained from segmentation as a conditional signal to guide the in-between process.

\textbf{Mask-conditioned Diffusion In-betweening.} 
Given a reference motion sequence $\boldsymbol{x} \in \mathbb{R}^{N \times D}$, the segmented human mask (obtained from SAM \cite{kirillov2023segany}) $\boldsymbol{m} \in \mathbb{R}^{N \times w \times h}$, and a keyframe signal $\boldsymbol{c} \in \mathbb{R}^{N}$ (mismatch detection results to identifies flaw motion), this module correct the reference motion by replacing mismatched motion frames.
We employ a pre-trained Vision Transformer (ViT) as the mask feature extractor to capture rich human pose information from the segmentation mask.
The mask combined with the motion context is used as the condition of the motion diffusion model.
Following \cite{karunratanakul2023gmd,cohan2024flexible}, we concatenate the resulting sample, keyframe signal, and mask features as model input to inform the generation model with condition signal.

\textbf{Training process.}
A random motion segment, selected at a random sequence position, is chosen as the generation target. Our model is trained to reconstruct this segment. Both keyframe conditioning signals $\boldsymbol{c}$ and mask conditioning signal $\boldsymbol{m}$ are set to $\emptyset$ for 10\% of training data to make our model suited for unconditioned motion generation.


\subsection{Physics-based Motion Transfer Module}

Our PTM consists of a pre-trained imitation controller and a test time adaptation strategy. The controller, as defined in the Motion Imitation section, obtains basic motion imitation ability with substantial motion prior. The TTA strategy (with a set of adaptation settings) is carefully designed to adapt the controller to explore a single specific motion, solving the dynamic modeling of noised high-difficulty motion.
Compared to previous methods, the pretrain and adapt pattern avoids limiting the model to a specific data domain. It models the dynamics for a single motion based on the normal prior, significantly enhancing the ability to handle out-of-domain and high-difficulty motions.

\textbf{Imitation Controller Pre-training.} 
Acquiring as much motion prior knowledge as possible during the pre-training phase is essential to accelerate the adaptation process. Therefore, we perform pre-training on 4 datasets: AMASS \cite{mahmood2019amass}, Human3.6M \cite{ionescu2013human3}, AIST++ \cite{li2021learn,aist-dance-db}, and Motion-X \cite{lin2023motionx} kungfu subset. The pre-training phase employs strict reconstruction rewards and early termination conditions.


\textbf{RL-based Test Time Adaptation.}
With a pre-trained imitation controller, rich motion priors open up the possibility for rapid adaptation during test time. Utilizing the trial-and-error nature of reinforcement learning, we propose RL-based test time adaptation, which involves performing a limited number of experiment steps on the current test data under specific adaptation settings (updating network parameters). Each motion sequence is treated as an individual instance, and the adaptation is performed for each instance independently. This method is particularly useful when dealing with high-difficulty and low-quality motions. The following designs are used in the adaptation process.

\textbf{Relative Reward.}
The captured reference motion contains jitter or fault roots with error accumulation, making constructing a full reconstruction reward detrimental.
Therefore, in the adaptation process, we designed a relative reward $r_{t}^{\mathrm{g}}$ that neglects the absolute root position, maintaining global orientation and translation through explicit guidance from rotation and implicit guidance from velocity.
The relative reward is formulated as: 
\begin{equation}
\begin{aligned}&r_{t}^{\mathrm{g}}=e^{w_{\mathrm{p}}\|rela(\boldsymbol{\hat{p}}_{t})-rela(\boldsymbol{p}_{t})\|}+e^{w_{\mathrm{r}}\|\boldsymbol{\hat{\theta}}_{t}\ominus\boldsymbol{\theta}_{t}\|}\\&+e^{w_{\mathrm{v}}\|\boldsymbol{\hat{v}}_{t}-\boldsymbol{v}_{t}\|}+e^{w_{\mathrm{\omega}}\|\boldsymbol{\hat{\omega}}_{t}-\boldsymbol{\omega}_{t}\|},\end{aligned}
\end{equation}
where $\boldsymbol{\hat{p}}_{t}$ means the joint position of reference motion, $rela()$ means to ignore the gravity axis part of root joints. $\ominus$ means rotation difference and $w$ is weights factor.


\textbf{Early Termination.} In high-difficulty motion tracking, there is a large displacement between humanoid and reference motion, as reference motion often involves frequent floating and penetration. This phenomenon makes defining when an adaptation step should be terminated challenging, which is crucial for adaptation efficiency and preventing undesirable behaviors. Therefore, we design a relative termination condition by calculating each joint's mean relative distance between humanoid and reference motion.
One adaptation step will be terminated when the distance exceeds threshold $d_{term}$. We also introduce termination condition $\mathcal{F}^h_t$ and $\mathcal{F}^c_t$ based on joint height and ground contact to consider falls and erroneous contacts occur. The full termination $\mathcal{F}_t$ is defined below, a smaller threshold $d_{term}$ indicating a stricter adherence to reference motion.
\begin{equation}
    \mathcal{F}_t = \left( \frac{1}{J} \sum_{i=1}^{J} \|rela(\boldsymbol{\hat{p}}^{i}_{t})-rela(\boldsymbol{p}^{i}_{t})\| > d_{term} \right) \vee \mathcal{F}^h_t \vee \mathcal{F}^c_t
\end{equation} 

\textbf{Residual Force.} During the test-time adaptation phase, we introduce residual forces \cite{yuan2020residual} to compensate for the dynamics mismatch.
This is important because complex motions often involve airborne flips and jumps (commonly seen in gymnastics and martial arts), which frequently rely on elastic trampolines and mats for execution. The use of external forces is necessary to account for the absence of these environmental conditions in our simulations.


\section{Experimental Results and Analysis}

\label{sec:expriment}

\subsection{Datasets}
We use four datasets to {train} our model: AMASS \cite{mahmood2019amass}, Human3.6M \cite{ionescu2013human3}, AIST++ \cite{li2021learn,aist-dance-db}, and Motion-X \cite{lin2023motionx} kungfu subset. 
AIST++ contains 5 hours of diverse dance motions, Motion-X is a huge motion generation dataset and its kungfu subset contains complex kungfu motions over 1k clips.
We perform our {evaluations} on the test set of AIST++, EMDB \cite{kaufmann2023emdb}, and kungfu. 
Sequences involving human-object interactions are removed for all datasets.

We collected 206 high-difficulty motion videos from YouTube, including rhythmic gymnastics (floor, ball, ribbon), dance (breakdancing, ballet, yoga), and martial arts (kungfu, taekwondo). We use these videos as in-the-wild data for evaluation. Compared to the previously mentioned datasets, these videos contain more complex motions, posing greater challenges for physics-based motion repair tasks. These data can also be used to evaluate the generalization capabilities of 3D human motion recovery methods, and we will make them publicly available.

\subsection{Metrics}
Following the latest method \cite{wang2025tram,shin2024wham,shen2024world}, we evaluate camera-coordinate metrics using the widely used MPJPE, Procrustes-aligned MPJPE (PA-MPJPE), Per Vertex Error (PVE), and Acceleration error (Accel). For world-coordinate metrics, we divide the global sequences into shorter segments of 100 frames aligning each segment with GT like GVHMR \cite{shen2024world}. We then report the World-aligned Mean Per Joint Position Error (WA-MPJPE$_{100}$), the World MPJPE (W-MPJPE$_{100}$), and the whole sequence for Root Translation Error (RTE, in \%).
In addition, we designed a benchmark to assess physical realism and motion reconstruction fidelity. This evaluation metric does not require 3D annotated data and is suitable for reflecting the model's generalization capability on in-the-wild motions.

\textbf{Physical realism.}
\textbf{1) Self-Penetration (SP)} measures self-intersection severity.
\textbf{2) Ground-Penetration (GP)} measures ground penetration
\textbf{3) Float} measures meshes floating above the plane. 
\textbf{4) Foot-Skate (FS)} measures foot sliding, we find feet that contact the ground in adjacent frames and calculate their average horizontal differences.

\textbf{2D Similarity.}
We utilize object segmentation and 2D keypoint detection methods to annotate our in-the-wild test set and design metrics for 2D and 3D Similarity.
\textbf{1) 2D Keypoint OKS.} We project the 3D motion onto 2D space and compute the Object Keypoint similarity with the 2D keypoints; a higher similarity indicates better restoration of the estimated 3D motion.
\textbf{2) Mask-Pose Similarity (MPS).} We project the 3D human mesh into the 2D camera plane and calculate the ratio of mesh points that fall within the segmented human mask. A larger ratio signifies higher motion restoration and a more accurate human shape estimation.

\subsection{Implementation details}
It takes around 2-3 days to get our pre-trained PTM with a single NVIDIA A100 GPU. During inference, restoring normal motions (such as running and jumping) requires fewer adaptation steps (less than 500) or may not require any adaptation at all. In contrast, restoring high-difficulty motions (such as continuous rolls and aerial maneuvers) necessitates between 2,000 and 4,000 steps, depending on the complexity of the motion and the quality of the reference action. We adopt the motion diffusion model GMD with UNet architecture as our in-between baseline \cite{dhariwal2021diffusion,ho2020denoising}. For more implementation details, please refer to the Appendix.

\begin{table*}
\begin{center}
\vspace{-2mm}
\scalebox{0.65}{
\begin{tabular}{lcccccccccccccc}
\toprule
\multirow{2}*{Datasets} & \multirow{2}*{Method} & \multicolumn{3}{c}{World Coordinate} & \multicolumn{4}{c}{Camera Coordinate} & \multicolumn{2}{c}{2D Similarity} & \multicolumn{4}{c}{Physical Authenticity}\\
\cmidrule(r){3-5} \cmidrule(r){6-9} \cmidrule(r){10-11} \cmidrule(r){12-15}
 &  & WA-MJE $\downarrow$& W-MJE $\downarrow$& RTE $\downarrow$& MPJPE $\downarrow$& PA-MPJPE $\downarrow$& PVE $\downarrow$& Accel $\downarrow$& OKS $\uparrow$& MPS $\uparrow$ & SP $\downarrow$ & GP $\downarrow$ & Float $\downarrow$ & FS $\downarrow$\\
\midrule
\multirow{7}*{AIST++}
&PhysPT \cite{zhang2024physpt} CVPR'24 & 139.974 & 218.344  & 9.307 & 97.143 & 68.026 & 115.007 & 8.406 &0.932&0.778& -- & 7.677 & 21.348 & 2.432\\ 
&TRAM \cite{wang2025tram} ECCV'25 & \underline{106.197} & \underline{159.520} & 9.433 & \textbf{91.809} & \textbf{64.024} & \textbf{107.334}& 7.727 & 0.945 & 0.786 & 0.150 & 20.557 & 489.984 & 2.350\\ 
&TRAM+PhysPT & 136.828 & 218.335 & 6.510 & 93.570 & 67.657 & 110.989 & 8.601 &0.903&0.757& -- & 4.079 & 22.688 & 2.066\\ 
&TRAM+Ours & \textbf{106.151} & \textbf{157.724} & 8.962 & 93.954 & {67.009} &  \underline{110.245} & 8.383 & 0.953 & 0.787 & \underline{0.046} & \underline{0.499} & \textbf{1.974} & \textbf{0.586}\\ 
\cmidrule(r){2-15}
&\small{GVHMR \cite{shen2024world} SIGGRAPH Asia'24} & 124.434 & 197.287 & \underline{5.083} & \underline{93.548} & \underline{65.245} & 111.548 & \textbf{6.850} & \textbf{0.965} & \underline{0.790} & 0.072 & 12.390 & 71.190 & 2.232\\ 
&GVHMR+PhysPT & 182.120 & 281.093 & 6.760 & 143.612 & 78.791 & 169.827 & 8.601 &0.905&0.764& -- & 4.978 & 27.052 & 2.468\\ 
&GVHMR+Ours & 123.365 & 193.792& \textbf{4.850} & 94.037 & 67.075 & 112.215 & \underline{7.302} & \underline{0.963} & \textbf{0.806} &\textbf{0.046} & \textbf{0.498} & \underline{1.982} & \underline{0.587}\\ 
\midrule
\multirow{7}*{Kungfu}
&PhysPT \cite{zhang2024physpt} CVPR'24 & 135.652 & 217.131 & 7.907 & 128.553 & 57.458  & 124.852 & 12.162 &0.919&0.765& -- & 23.630 & 94.647 & 10.955\\ 
&TRAM \cite{wang2025tram} ECCV'25 & 113.354 & 209.664 & 7.539 &\underline{}{84.610} &\underline{55.735}  & \underline{101.079} & \underline{11.872} &0.925&0.761& 0.136 & 4.320 & 40.924 & 2.574\\ 
&TRAM+PhysPT & 174.394 & 344.192 & 7.752 & 119.675 &60.467  & 141.917 & 12.912 &0.916&0.713& -- & 3.193 & 21.653 & 1.146\\ 
&TRAM+Ours & 113.284 &\textbf{193.673} & 7.211& \textbf{}{79.493} &{57.714}  & \textbf{91.636} & 12.626 &0.931&\underline{0.775}& \underline{0.077} & \textbf{0.239} & \textbf{5.714} & \underline{0.259}\\ 
\cmidrule(r){2-15}
&\small{GVHMR \cite{shen2024world} SIGGRAPH Asia'24} & \textbf{106.763} & 204.495 & \underline{4.868} & 96.316 & \textbf{54.748} & 113.218 & \textbf{11.630} &\textbf{0.958}&0.765& 0.079 & 10.368  & 43.401 & 2.217\\ 
&GVHMR+PhysPT & 211.972 & 344.590 & 8.605 & 97.270 &55.923  & 112.178 & 14.988 &0.902&0.696& -- & 3.189 & 26.097 & 1.774\\ 
&GVHMR+Ours & \underline{109.986} & \underline{198.090} & \textbf{4.699} & 97.995 & 57.061 & 112.868 & 12.234 &\underline{0.953}&\textbf{0.792}& \textbf{0.018} & \underline{0.290}  & \underline{6.240} & \textbf{0.257}\\ 
\midrule
\multirow{7}*{EMDB}
&PhysPT  \cite{zhang2024physpt} CVPR'24    & 285.464 & 741.967 & 10.838 & 264.547 &40.952  & 307.372 & 5.9063 & \textbf{0.956} & 0.793& -- & 1.855 & 21.144 & 2.738\\
&TRAM    \cite{wang2025tram} ECCV'25    & 230.633 & 322.495 & 3.162 & 266.600 &38.474  & 305.433 & \textbf{5.546} & 0.947& 0.792 & 0.073 & 199.710 & 161.200 & 17.373\\
&TRAM+PhysPT & 358.803 & 881.275 & 11.627 & 256.744 &40.619  & \underline{298.817} & 6.791 &0.908&0.767& -- & 2.382 & 11.686 & 1.985\\ 
&TRAM+Ours   & 224.038 & 312.028 & 2.074 & 263.955 & 40.396 & \textbf{295.395} & 6.194 & 0.949& 0.797& 0.045 & \underline{1.415} & \underline{4.562} & {1.482}\\
\cmidrule(r){2-15}
&\small{GVHMR \cite{shen2024world} SIGGRAPH Asia'24}      & \underline{109.104} & \underline{274.941} & \underline{1.960} & {252.159} & \underline{38.112} &316.509  & \underline{5.870}  & \underline{0.954} & \underline{0.801} & \underline{0.006} & 82.266 & 510.298 & 0.693\\ 
&GVHMR+PhysPT   & 781.128 & 1491.893 & 14.588 & \underline{251.277} &50.333  & 303.236 & 6.652 &0.916&0.751& -- & 0.983 & 9.924 & \underline{0.494}\\ 
&GVHMR+Ours     & \textbf{91.153} &  \textbf{261.579}  &\textbf{1.180} &\textbf{249.127} &\textbf{37.897}&313.925  & 6.694 &0.948&\textbf{0.807}& \textbf{0.002} & \textbf{0.248}  & \textbf{3.632} & \textbf{0.173}\\ 
\bottomrule
\end{tabular}
}
\end{center}
\vspace{-4mm}
\caption{\textbf{Evaluation on multiple video motion capture dataset.} Since our method is based on physical simulation, we filtered these datasets and removed the human-object interaction scenes. WA-MJE and W-MJE mean WA-MPJPE$_{100}$ and W-MPJPE$_{100}$ separately.}
\vspace{-3mm}
\label{table:Datasets}
\end{table*}

\begin{table}
\begin{center}
\scalebox{0.7}{
\begin{tabular}{ccccccc}
\toprule
Method  & OKS $\uparrow$ & MPS $\uparrow$ & SP $\downarrow$ & GP $\downarrow$ & Float $\downarrow$ & FS $\downarrow$ \\
\midrule
PhysPT & 0.687 & 0.497 & -- &4.789 & 38.189  & 4.436 \\
TRAM & 0.828 & 0.667 &0.438 &19.988 &107.432&12.261\\
TRAM+PhysPT & 0.730 & 0.645 &-- & 7.883 &39.379&6.007\\
TRAM+Ours & {0.845} & {0.687} &0.363 &0.595 &16.956  & 0.779\\ 
\midrule
GVHMR & 0.837 & 0.704 &0.289 &9.999 &137.969  & 3.006 \\
GVHMR+PhysPT & 0.806 & 0.685 &-- &6.616 &54.032&5.630\\
GVHMR+Ours & \textbf{0.854} & \textbf{0.710} &\textbf{0.120} &\textbf{0.334} &\textbf{14.921}  & \textbf{0.717} \\
\bottomrule
\end{tabular}
}
\end{center}
\vspace{-5mm}
\caption{\textbf{Evaluation of our collected in-the-wild dataset.}}
\vspace{-2mm}
\label{table:youtube}
\end{table}

\begin{table}
\begin{center}
\scalebox{0.7}{
\begin{tabular}{cccccc}
\toprule
Method  & SR $\uparrow$ &MPJPE$_g$ $\downarrow$ & MPJPE $\downarrow$ & MPJPE$_{pa}$ $\downarrow$ \\
\midrule
UHC & 42.91\% & 86.23 & 48.91 & 39.73 &   \\
PHC+ & 76.41\%  &84.86 & 47.98 & 39.43 &  \\
PTM & \textbf{98.16\%} & \textbf{82.13} & \textbf{33.45} & \textbf{26.12} & \\ 
\bottomrule
\end{tabular}
}
\end{center}
\vspace{-5mm}
\caption{\textbf{Physical transfer ability.} The reference motions are from the kungfu subset, following PHC+ in metrics calculation.}
\vspace{-2mm}
\label{table:imitation}
\end{table}

\begin{table}
\begin{center}
\scalebox{0.85}{
\footnotesize
\begin{tabular}{ccccccc}
\toprule
Early-Term &  Res-F& TTA &  Rela-Rwd  & OKS $\uparrow$ & MPS $\uparrow$ & SR $\uparrow$\\
\midrule
& & &  & 0.811 & 0.673 & 37\%\\ 
\checkmark& & &  & 0.784 & 0.652 & 52\%\\ 
\checkmark &  \checkmark & &   & 0.823 & 0.673 & 61\%\\ 
\checkmark & \checkmark & \checkmark &  & 0.850 & 0.706 & 85\%\\ 
\checkmark & \checkmark & \checkmark & \checkmark  & \textbf{0.853} & \textbf{0.710} &\textbf{87\%}\\ 
\bottomrule
\end{tabular}
}
\vspace{-3mm}
\caption{\textbf{Effectiveness of adaptation settings in our PTM.}}
\vspace{-4mm}
\label{table:RLTTA_ab}
\end{center}
\end{table}

\begin{table}
\begin{center}
\scalebox{0.85}{
\footnotesize
\begin{tabular}{cccccccc}
\toprule
\multirow{2}*{In-between} & \multicolumn{2}{c}{Condition} & \multicolumn{2}{c}{Match-Detect} & \multicolumn{3}{c}{Metrics}\\
\cmidrule(r){2-3} \cmidrule(r){4-5} \cmidrule(r){6-8}
 & mask & kpts & mask& kpts & OKS $\uparrow$ & MPS $\uparrow$ & SR $\uparrow$\\
\midrule
& & & & & 0.802 & 0.677 & 78\%\\ 
\checkmark& & & & \checkmark & 0.834 & 0.699&83\%\\ 
\checkmark& & & \checkmark & & 0.827 & 0.704 & 85\%\\ 
\checkmark& &\checkmark &\checkmark & & 0.845 & 0.706 & \textbf{87\%}\\ 
\checkmark&\checkmark & & \checkmark& & \textbf{0.853} & \textbf{0.710} & \textbf{87\%}\\ 
\bottomrule
\end{tabular}
}
\vspace{-2mm}
\caption{\textbf{Comparison of different motion correction settings.} In-between means whether to use the in-between module to correct the bad frames. The condition tells which signal we used to guide the diffusion in-between process. Match-Detect shows which mismatch detection method is more effective.}
\label{table:inbetween_ab}
\vspace{-5mm}
\end{center}
\end{table}


\begin{figure*}[t]
    \centering
\includegraphics[width=\textwidth]{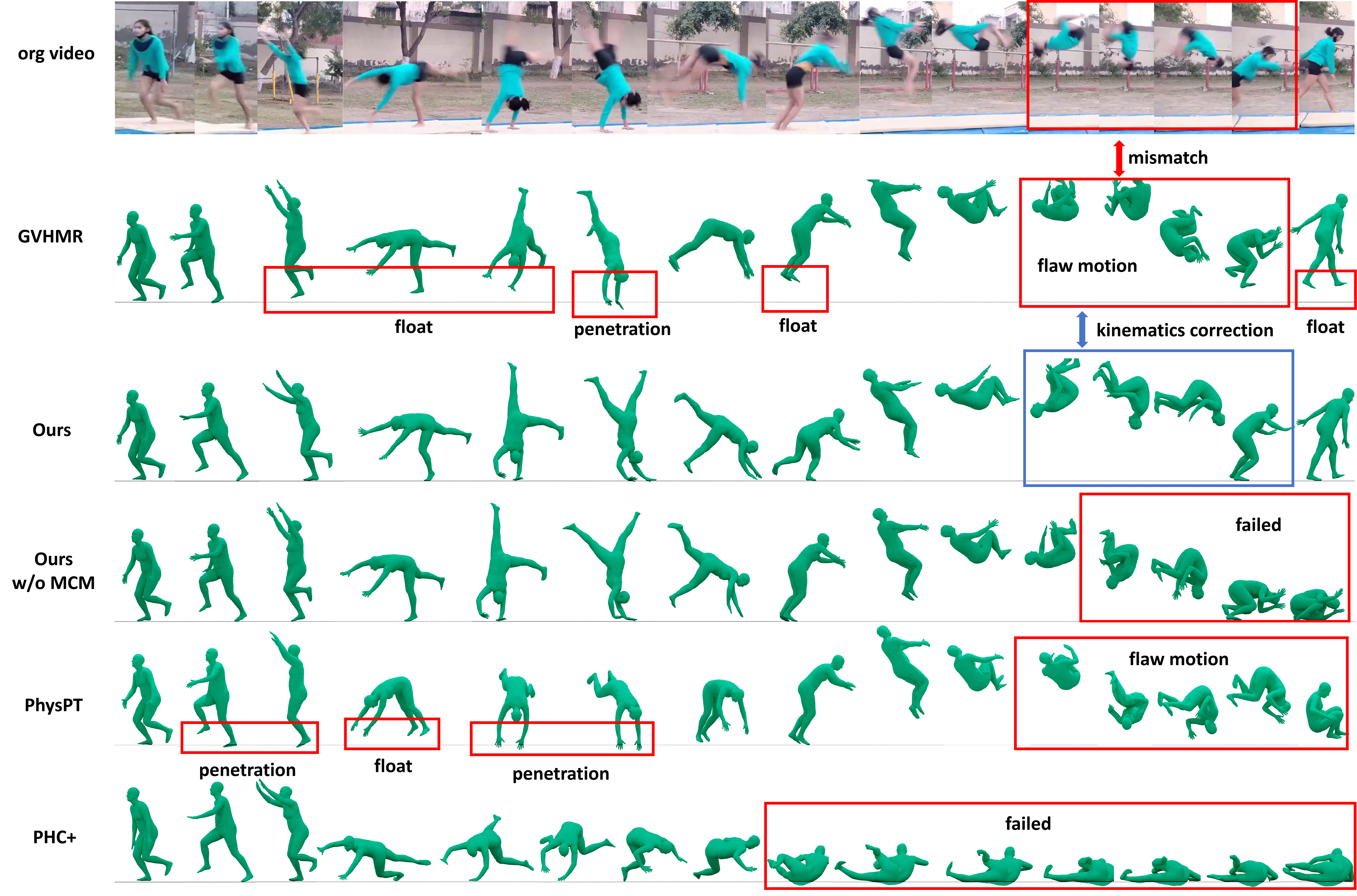}
    \vspace{-4mm}
    \caption{\label{fig:compare}\textbf{Qualitative comparison with state-of-the-art method.} }
     \vspace{-5mm}
\end{figure*}

\subsection{Comparison with the State-of-the-Art}
We selected two state-of-the-art (SOTA) video motion capture methods, TRAM \cite{wang2025tram} and GVHMR \cite{shen2024world}, and SOTA physical informed method PhysPT \cite{zhang2024physpt} for comparison. The comparison results are presented in Table \ref{table:Datasets}. 

For \textbf{world coordinate} metrics, our method outperforms the original motions in most cases.
This improvement stems from the direct relationship between the world coordinate system and physical space. 
Particularly in the EMDB dataset, where prolonged displacement leads to the accumulation of errors in local perspectives over time and space, our method effectively mitigates these issues, resulting in improvements in world coordinate metrics.
Previous simulation methods typically emphasized perfect tracking of reference motions, focusing more on imitation rather than restoration. Consequently, when there is a significant deviation in the root position along the gravity axis, these methods often fail due to the influence of gravity, especially when attempting to imitate complex or flawed motions. In contrast, our approach prioritizes repair, utilizing a unique PTM that enables stable tracking of high-difficulty motions.

For 3D motion restoration in the \textbf{camera coordinate}, our method shows minor discrepancies from the original motions in both the Kungfu and AIST++ datasets. This can be attributed to two main reasons:
1) Our method uses noisy motions as input without knowledge of the GT motions, which inherently disadvantages us.
2) Our repair approach models directly in physical space without considering camera parameters, making it challenging to achieve perfect restoration from specific camera viewpoints.
Despite these challenges, our results on the EMDB dataset exhibit slight improvements compared to the original motions. This is because our method accounts for the physical space and offers a better understanding of long-term global trajectory movements, whereas noisy motions accumulate errors due to changes in time and spatial position.

Regarding \textbf{2D similarity}, the repaired motions in the Kungfu dataset show slight improvements over the original motions. This is due to current video motion capture methods being prone to brief flaw motions when dealing with complex motions, which become more pronounced when projected onto 2D images. Our MCM replaces flaw motion based on the human masks and motion context, enhancing the 2D restoration of the repaired motions.

In terms of \textbf{physical authenticity} metrics, our method successfully restored the physical realism of motions, resulting in significant improvements in ground penetration, foot sliding, and floating. 
Our method keeps the ground penetration for all datasets below 0.5; notably, for the EMDB dataset, we reduced the ground penetration from as high as 82 to 0.24. This is attributed to the long-term global trajectory changes, where the errors in the gravity axis accumulate along movements, leading to severe ground penetration and floating. 
For all three datasets, our method reduces the self-penetration by over 50\%. In the Kungfu dataset, we successfully lowered the self-penetration of GVHMR from 0.079 to 0.018. Furthermore, foot sliding also shows consistent improvement for all datasets, largely due to the presence of friction in the physical environment. While motion simulation offers inherent advantages, our differentiation from other physics-based simulation methods lies in the ability to track in-the-wild and complex motions. 

%

\subsection{Qualitative Evaluation}
In Figure \ref{fig:compare}, we select high-difficulty in-the-wild motions for visualization and illustrate a comparison against SOTA techniques. GVHMR captures human motions from video and acts as a noise motion generator for PhysPT, PHC+ \cite{luo2023universal}, and our methods.
GVHMR successfully captures human motion from a monocular camera, yet the resulting motion exhibits significant physical issues such as floating and penetration. Additionally, it manifests short-term flawed motion when faced with more complex movements. 
Due to simplified physical rules and the unawareness of the high-difficulty motion distribution, PhysPT faces challenges in both physical repair and preservation of the original motion when dealing with complex motions. Moreover, it is ineffective in addressing flawed motions. 
The advanced motion imitation method PHC+ is capable of tracking large-scale motion capture datasets but fails on high-difficulty noisy motions. This is attributed to PHC+'s lack of generalization ability for complex movements and its susceptibility to noise in reference motions.
In contrast, despite the original motions containing multiple complex motions that are challenging to reproduce in physical space, our method eliminates ground penetration and floating issues while successfully maintaining the original motion patterns.
Furthermore, we implement MCM to correct the flawed motion, avoiding the simulation failures caused by flawed motion and ensuring harmony with the video and motion context.




\subsection{Ablation Studies}

\textbf{Physical Transfer Ability.}
We evaluate the simulation capabilities of our proposed PTM using the Kungfu dataset, comparing them against SOTA motion imitation methods, the results are presented in Table \ref{table:imitation}. 
Compared to outstanding motion imitation methods UHC and PHC+, our PTM demonstrates commendable performance in tracking martial arts, achieving a 98\% success rate and considerable enhancements across other metrics.
These results affirm the superior capability of our methods in motion imitation, particularly in tracking high-difficulty motions.
The improvement in metrics is attributed to the relative design tailored for complex motions and the ability of TTA to mitigate the forgetting phenomenon. Previous studies have indicated that motion imitation can suffer from rapid loss of earlier knowledge when attempting to imitate a wide array of motions \cite{luo2023perpetual}. In our PTM, the controller's memory serves as a cornerstone for learning new actions, allowing the optimization of specific data samples without retaining this memory. We aim for the model to proactively explore solutions rather than merely reproducing answers it has memorized.

\textbf{Effect of RL-TTA Strategy.}
In Table \ref{table:RLTTA_ab}, we conduct an ablation study on various components of the RL-TTA strategy, with experiments carried out on a high-difficulty in-the-wild dataset to demonstrate its effectiveness in handling challenging movements. 
While the pre-trained controller can track daily motions well, it struggles when facing difficult movements.
The reason why success rates increase with early termination is that traditional early termination strategies impose overly strict requirements on humanoids, making it easy to fail when facing poor-quality motion, greatly hindering the learning process.
Since high-difficulty motion often involves airborne motions that rely on external apparatuses such as trampolines, the absence of residual forces prevents humanoids from executing these movements. The TTA contributes the most enhancement by simplifying the humanoid's objectives to a specific action.

\textbf{Effect of Mask-conditioned Motion Correction.}
As shown in Table \ref{table:inbetween_ab}, we conduct an ablation study on our MCM, which corrects flawed motions in reference motion. The experiments are performed on a high-difficulty in-the-wild dataset. 
Experimental results show that our MCM can enhance the simulation success rate and improve 2D similarity by replacing flaw motions that result in simulation failure. 
Regarding modality selection, we find that human masks outperform 2D keypoints, as masks contain more comprehensive shape and motion information.
When facing complex movements, keypoint detection algorithms may misidentify or overlook some joints, while segmentation algorithms exhibit greater stability and only require distinguishing between human foreground and background.

\section{Conclusion}
This paper introduces a plug-and-play motion restoration method to enhance the physical quality of in-the-wild high-difficulty motions. Our method integrates easily with any video motion capture method, greatly improving the efficiency of obtaining high-quality 3D motions. The MCM accurately corrected the flawed motion in video motion capture results, while the PTM successfully achieved the physical restoration of high-difficulty in-the-wild motions. Comprehensive experiments showcase our model's performance and highlight each module's contributions and impacts. 
Our work provides valuable insights for future research in this field.
The main limitation of our work is that it can only handle single-person motions and is unable to restore closely interactive multi-person movements.


{
    \small
    \bibliographystyle{ieeenat_fullname}
    \bibliography{main}
}
\end{document}